\pdfoutput=1

\documentclass[11pt]{article}

\usepackage[final]{acl}

\usepackage{times}
\usepackage{latexsym}
\usepackage{booktabs}
\usepackage{multirow}
\usepackage{comment}
\usepackage[breakable]{tcolorbox}
\usepackage{subcaption}
\usepackage{colortbl}
\usepackage{hyperref}

\usepackage[T1]{fontenc}

\usepackage[utf8]{inputenc}

\usepackage{microtype}

\usepackage{inconsolata}

\usepackage{graphicx}

\newif\ifcomments
\commentstrue
\ifcomments
    
\fi

\title{Good Intentions Beyond ACL: Who Does NLP for Social Good, and Where?}

\author{
  \textbf{Grace LeFevre\textsuperscript{1}},
  \textbf{Qingcheng Zeng\textsuperscript{1}},
  \textbf{Adam Leif\textsuperscript{2}},
  \textbf{Jason Jewell\textsuperscript{1}},
  \textbf{Denis Peskoff\textsuperscript{1}},
  \textbf{Rob Voigt\textsuperscript{3}}
\\
  \textsuperscript{1}Northwestern University,
  \textsuperscript{2}University of California, Los Angeles,\\
  \textsuperscript{3}University of California, Davis
\\
  \texttt{
    gracelefevre@u.northwestern.edu, 
    robvoigt@ucdavis.edu
  }
}

\begin{document}
\maketitle
\begin{abstract}
The social impact of Natural Language Processing (NLP) is increasingly important, with a rising community focus on initiatives related to NLP for Social Good (NLP4SG).
Indeed, in recent years, almost 20\% of all papers in the \textsc{acl} Anthology address topics related to social good as defined by the UN Sustainable Development Goals \cite{adauto-etal-2023-beyond}. 
%
In this study, we take an author- and venue-level perspective to map the landscape of NLP4SG, quantifying the proportion of work addressing social good concerns both within and beyond the \textsc{acl} community, by both core \textsc{acl} contributors and non-\textsc{acl} authors. 
With this approach we discover two surprising facts about the landscape of NLP4SG. 
First, \textsc{acl} authors are dramatically more likely to do work addressing social good concerns when publishing in venues outside of \textsc{acl}. 
Second, the vast majority of publications using NLP techniques to address concerns of social good are done by non-\textsc{acl} authors in venues outside of \textsc{acl}. 
We discuss the implications of these findings on agenda-setting considerations for the \textsc{acl} community related to NLP4SG.

%
%
%
%

\end{abstract}

\section{ACL and Social Good Research}
\label{sec:socialgood}

\begin{figure}[t!]
    \centering
    \includegraphics[width=\linewidth]{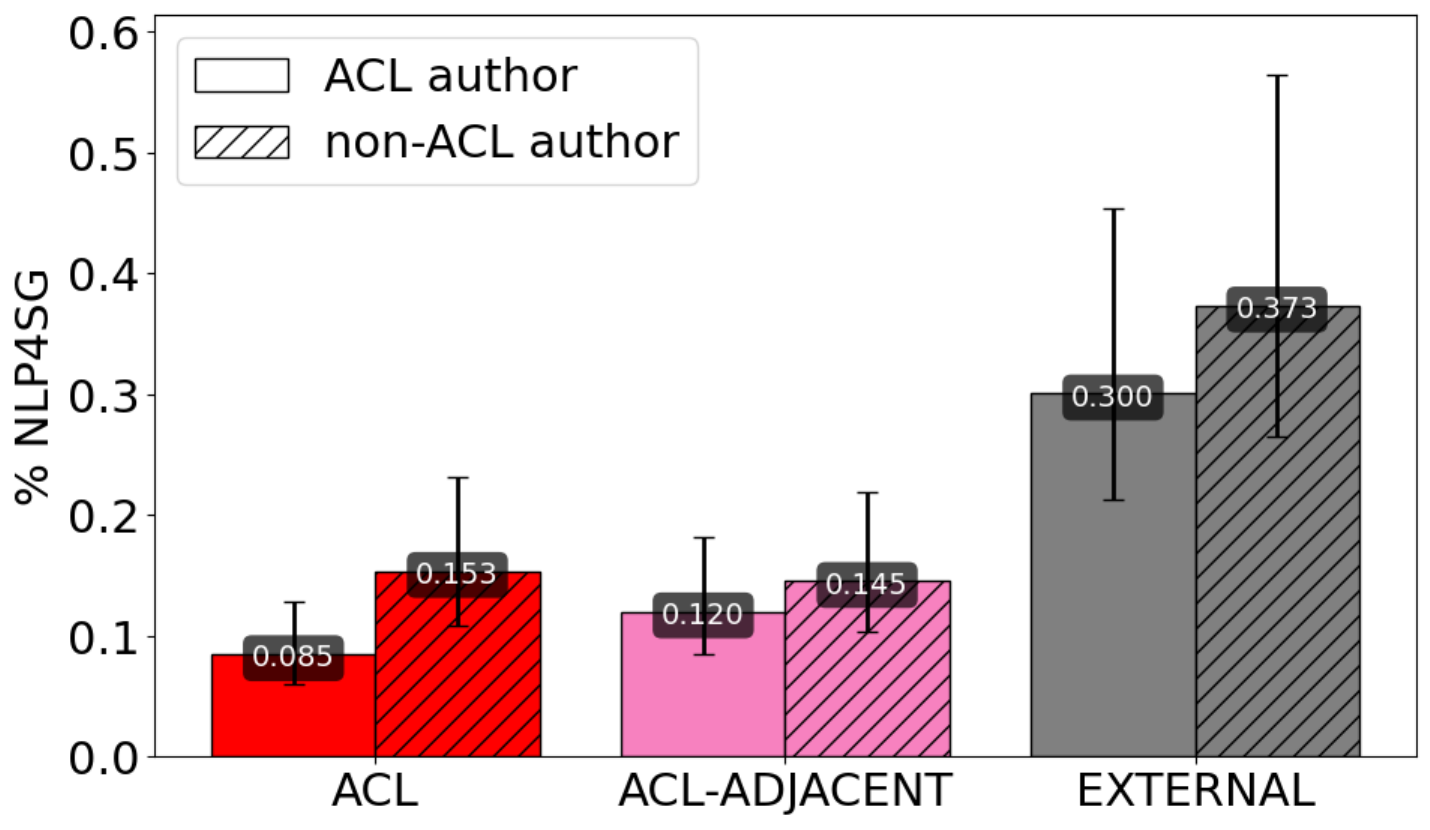}
    \caption{A higher ratio of NLP papers outside of the ACL Anthology (\textsc{external}) are characterized as social good than in \textsc{acl} and \textsc{acl-adjacent} venues. Moreover, NLP papers by non-\textsc{acl} authors are more likely to focus on social good questions than those by \textsc{acl} authors across all venue types.}
    \label{fig:f1}
\end{figure}

As natural language processing rises in prominence throughout society, ``NLP for Social Good” ~\citep[NLP4SG]{jin-etal-2021-good} is an increasingly important topic of conversation in the NLP community: how can NLP methods be used to address questions of social concern and applied for positive social impact?
Existing research on the \textsc{acl} Anthology shows that while a substantial minority of papers have addressed social good since the 1980s, the proportion has increased over time. Under the definition of papers that address questions relevant to one of the 17 United Nations Sustainable Development Goals (SDGs),\footnote{\citet{gosselink2024ai} and \citet{karamolegkou2025nlp} also map NLP4SG work onto the UN SDGs.}~\citet{adauto-etal-2023-beyond} find that the proportion of papers in the \textsc{acl} community addressing social good concerns has approached 20\% in recent years.

However, many venues beyond \textsc{acl} publish NLP4SG work. The proliferation of NLP techniques in other fields has led to substantial growth of papers that both use NLP methods and tackle social good questions, across dozens of conferences outside of the \textsc{acl} Anthology \citep{movva2024topics}.
In this environment, critical questions of agenda-setting arise for the \textsc{acl} community: 
\begin{quote}
     (1) When \textsc{acl} authors are doing work oriented towards social good, do they send that work to \textsc{acl} venues? 

     (2) Is \textsc{acl} the place where most NLP4SG work is taking place? 
\end{quote}

\begin{figure*}[t!]
    \centering
    \includegraphics[width=\linewidth]{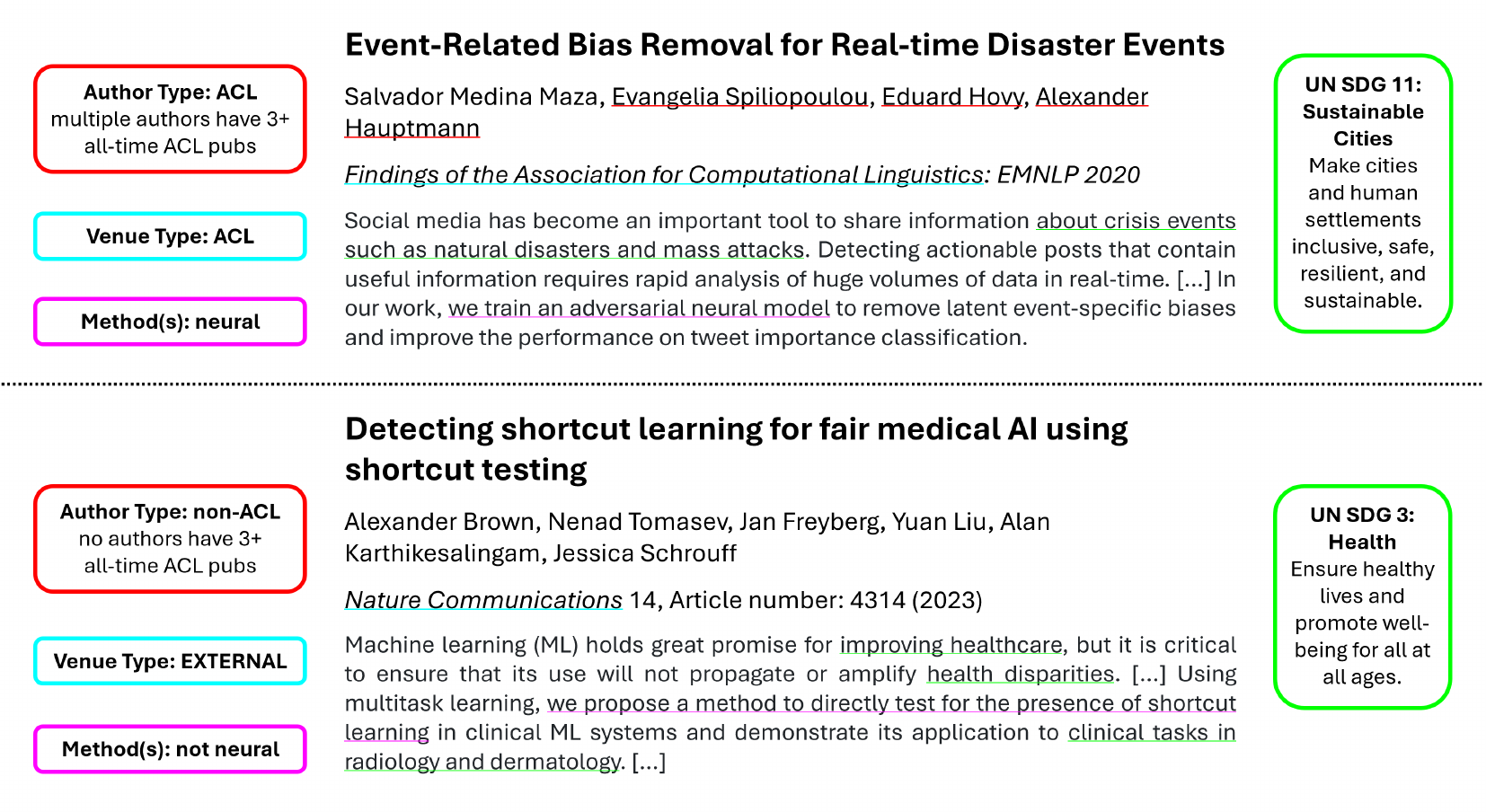}
    \caption{Schematic of the metadata augmentation used in conducting our analyses. Papers are labeled for relevance to social good as defined by UN SDGs, author association with ACL, venue type, and neural vs. non-neural methods.}
    \label{fig:descr_method}
\end{figure*}

Clear answers to these questions can help orient the community towards next steps moving forward. The first addresses whether \textsc{acl} authors themselves consider our community to be a welcoming and appropriate venue for NLP4SG work; the second, the centrality of our community and the broader reach of our methods towards often inherently interdisciplinary social good goals. 

In this work, we aim to gain a deeper understanding of the broader landscape of NLP4SG by adopting a scientometric approach that examines authors and publication venues \cite{https://doi.org/10.1002/asi.22630}.

We first augment an existing corpus of scientific papers with annotations for whether the work uses NLP techniques, whether the authors have published substantially within \textsc{acl}, and the venue type (\textsc{acl}, \textsc{acl-adjacent}, or \textsc{external}; Section~\ref{sec:methodology}).  

Analyzing these data, we find that papers by \textsc{acl} authors in venues outside of the \textsc{acl} Anthology are more than \textit{three times as likely} to address social good topics as those inside it, and that the substantial majority of NLP4SG work takes place beyond \textsc{acl} by non-\textsc{acl} authors.

Furthermore, we examine and find substantial topical differences in content across venue types, differences in the probability that work using NLP techniques addresses social good when segmenting venues by discipline, and venue- and author-based differences in the proportion of NLP4SG work relying on neural methods (Section~\ref{sec:distributional}).
We conclude by stressing the importance of defining social good and suggesting actions for closing the social good gap between \textsc{acl} and other venues~(Section~\ref{sec:conclusion}).

\section{Augmenting a Corpus with Metadata}

\label{sec:methodology}

This type of analysis requires a collection of relevant and representative academic papers.
\citet{adauto-etal-2023-beyond}, the work which we are most directly building upon, collect a dataset of 76,229 papers from \textsc{acl} Anthology and analyze a sample of 5,000 papers for social good, as defined by the United Nations Sustainable Development Goals (SDGs). 
We aim to expand beyond \textsc{acl} and use Semantic Scholar's Open Research Corpus (S2ORC) as a starting point~\citep{lo-etal-2020-s2orc}.
%

%
The metadata of the papers is paramount for our analysis.
While S2ORC maintains metadata in addition to full paper text, it is sourced from the paper text and due to nonspecific PDF text extraction, some is missing or incorrect.
We expand our data with the Semantic Scholar Open Data Platform \citep{kinney2023semantic} via the S2AG API, which stores additional metadata for each paper.\footnote{https://api.semanticscholar.org/api-docs/datasets} 
%
%
%
%
We further augment the corpus by linking each paper in S2ORC to a corresponding record in OpenAlex \citep{priem2022openalexfullyopenindexscholarly}, which improves the distinction of authors with similar names and provides concepts for grouping our Semantic Scholar corpus.\footnote{If a paper's MAG or DOI is available in the Semantic Scholar metadata, this is used to identify that same paper in OpenAlex; otherwise, papers are matched via their title and publication date/year (whichever is available).}
For \textsc{acl} papers, we also augment each paper's metadata with information from the \textsc{acl} Anthology BibTeX. Finally, we use the resulting corpus to identify four key factors for each paper: 
\begin{itemize}
\itemsep0em 
\item NLP relevance
\item venue type
\item author classification
\item social good relevance
\end{itemize}

The schematic in Figure \ref{fig:descr_method} illustrates our metadata augmentation process for two representative papers. We release the resulting dataset, along with the code used to create it, to the community to encourage future work.\footnote{\url{https://github.com/asl7168/nlp4sg_beyond_acl}}

\subsection{NLP Relevance}

OpenAlex provides a classification of relevant ``concepts'' for each paper in their database, which we leverage to categorize which papers involve NLP. We identify a set of core concepts associated with NLP by calculating the most frequently occurring OpenAlex concepts in \textsc{acl} Anthology papers and removing ambiguous or multi-sense concepts (like ``Computer Science''). 
Our full list of used concepts is in Table~\ref{tab:concepts}, 
including only clearly NLP-associated concepts like ``Natural Language Processing,'' ``Information Retrieval,'' and ``Language Model.''

Filtering our dataset to include only papers classified with at least one of these topics yielded our main dataset of 309,208 papers. 

\begin{table}[h!]
    \centering
    \begin{tabular}{ll} 
        \toprule
        \textbf{OpenAlex ID} & \textbf{Concept} \\
        \midrule
        C204321447 & Natural Language Processing \\
        C23123220 & Information Retrieval \\
        C203005215 & Machine Translation \\
        C119857082 & Machine Learning \\
        C186644900 & Parsing \\
        C28490314 & Speech Recognition \\
        C137293760 & Language Model \\
        \bottomrule
    \end{tabular}
    \caption{Concepts selected from OpenAlex.}
    \label{tab:concepts}
\end{table}

\subsection{Venue Type}
\label{subsection:venue_type}

We propose a distinction between three types of venues: 
\begin{itemize}
\itemsep0em 
\item \textsc{acl} venues are those listed as ``\textsc{acl} Events'' in the \textsc{acl} Anthology.
\item \textsc{acl-adjacent} venues are those listed as ``Non-\textsc{acl} Events'' in the Anthology.
\item \textsc{external} venues belong to neither list and include a diverse range of journals and conference proceedings from other disciplines, such as general science, engineering, and interdisciplinary research. The top 10 most frequent external venues in our dataset are listed in Appendix \ref{sec:external-venues}.
\end{itemize}

For \textsc{external} venues we further obtain coarse- and fine-grained disciplinary classifications, as well as venue-level h5-indices for a subset of \textsc{external} venues using the ``top publications'' metrics on Google Scholar.\footnote{\url{https://scholar.google.com/citations?view_op=top_venues&hl=en}} Specifically, for each of their eight top-level disciplinary categories, we scrape the top 20 venues under every subcategory and align them to our existing venue names. To account for variation in formatting (abbreviations, punctuation, etc.), we normalize venue names and apply a token-based fuzzy matching algorithm\footnote{We use the \texttt{RapidFuzz} library's \texttt{token\_sort\_ratio} scorer to align venue names based on string similarity, retaining only matches with a similarity score of 90 or higher.} that compares similarity scores and retains only high-confidence matches. This process results in 3,281 venues in this subset. These correspond to 98,753 papers, or 32\% of our original dataset, which are likely to be higher-impact and more reputable venues on average. We use this subset for detailed venue-level analysis  in Section \ref{section:venue_discipline}.

\begin{table}[t]
\centering
\begin{tabular}{llll}
\cline{2-4}
 & \multicolumn{3}{|c|}{\cellcolor{lightgray}{\textsc{venue}}} \\ \cline{1-4} 
\multicolumn{1}{|l|}{\cellcolor{lightgray}\textsc{author}} & \multicolumn{1}{c|}{\textbf{\textsc{acl}}} & \multicolumn{1}{c|}{\textbf{\textsc{acl-adj.}}} & \multicolumn{1}{c|}{\textbf{\textsc{external}}} \\ \hline
\multicolumn{1}{|l|}{\textit{ACL}} & \multicolumn{1}{c|}{24{,}594} & \multicolumn{1}{c|}{30{,}269} & \multicolumn{1}{c|}{28{,}644} \\ \hline
\multicolumn{1}{|l|}{\textit{non-ACL}} & \multicolumn{1}{c|}{1,075} & \multicolumn{1}{c|}{2,700} & \multicolumn{1}{c|}{221{,}926} \\ \hline
\end{tabular}
\caption{Distribution of papers in our dataset. Unsurprisingly, \textsc{acl} authors (those with three or more publications in an \textsc{acl} venue) are responsible for the majority of \textsc{acl} and \textsc{acl-adjacent} papers. Far more NLP papers exist in \textsc{external} and are primarily written by non-\textsc{acl} authors.}
\label{tab:data-distribution}
\end{table}

\subsection{Author Classification}

We propose an author-level distinction between ``\textsc{acl} Authors'' and ``non-\textsc{acl} Authors.'' We define \textsc{acl} authors as authors who, at any time, have published three or more papers in \textsc{acl} venues, and define any given paper as being written by an \textsc{acl} author if at least one author meets this criterion. This author-level distinction is important because it accounts for an author's experience within the \textsc{acl} community in addition to considering whether a paper appears in an \textsc{acl} venue. 

By defining \textsc{acl} authors based on their publication history, we aim to better differentiate between researchers with sustained engagement in the field and those with more limited exposure. Notably, our threshold of three \textsc{acl} papers aligns with \textsc{acl}’s own requirement for becoming a reviewer, reflecting a level of experience that the community considers sufficient for evaluating research.

The distribution of papers in our dataset across each author and venue type is shown in Table \ref{tab:data-distribution}.

\subsection{Social Good Relevance}

We apply the NLP4SG model from \citet{adauto-etal-2023-beyond} to classify each paper as relevant or not relevant to social good. However, applying this model to papers from outside the \textsc{acl} Anthology may have limited accuracy due to domain transfer issues. Therefore, we conduct a manual evaluation of the accuracy of this model on papers in non-\textsc{acl} venues. 
Three authors with backgrounds in NLP annotated 200 randomly selected papers published by an \textsc{acl} author at a non-\textsc{acl} venue for their association with social good, as defined by association with UN SDGs. Annotation instructions are provided in Appendix \ref{sec:annotation_instructions} and results in Table \ref{tab:domain_transfer}. 

Overall, the NLP4SG classifier aligns with the human judgment 77.5\% of the time, achieving a precision of 71.0\% and a recall of 66.2\% for an overall F1 score of 68.5\%. As expected for out-of-domain classification, this represents some loss of performance over \citet{adauto-etal-2023-beyond}, who reported an F1 score of 75.9\% for papers in the \textsc{acl} Anthology. 
However, we find the model is not dramatically unbalanced in its predictions. Due to lower recall, it is conservative at assigning NLP4SG labels to papers in \textsc{external} venues. Therefore, it is sufficient for the large-scale trends we track in this work and we use its labels at face value. To reflect the uncertainty introduced by the classifier's performance, we include error bars in Figure \ref{fig:f1}, estimating the lower bound as \( \mathit{value} \times \mathit{precision} \) and the upper bound as \( \mathit{value} \div \mathit{recall} \). These account for potential over- and under-estimation due to false positives and false negatives, respectively.

\begin{figure}[ht!]
    \centering
    \includegraphics[width=\linewidth]{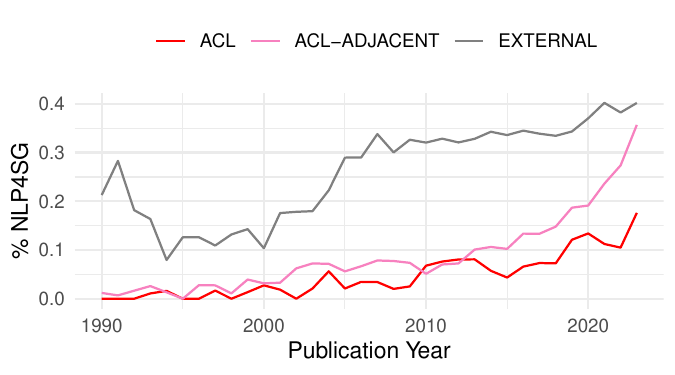}
    \caption{Proportion of NLP papers identified as NLP4SG by venue type and year. Applications of NLP techniques to social good questions are increasing as a share of all NLP papers across all venues.}
    \label{fig:yearly}
\end{figure}
\begin{table}[t]
    \centering
    \begin{tabular}{lcc}
    \cline{2-3}
        & \multicolumn{2}{|c|}{\cellcolor{lightgray}\textsc{human}} \\
    \cline{1-3}
      \multicolumn{1}{|c|}{\cellcolor{lightgray}\textsc{classifier}}  & \multicolumn{1}{c|}{NLP4SG} & \multicolumn{1}{c|}{Other} \\
      \cline{1-3}
       \multicolumn{1}{|l|}{\textit{NLP4SG}} &  \multicolumn{1}{c|}{49} & \multicolumn{1}{c|}{20}  \\
        \cline{1-3}
        \multicolumn{1}{|l|}{\textit{Other}} & \multicolumn{1}{c|}{25}  & \multicolumn{1}{c|}{106}  \\
        \cline{1-3}
    \end{tabular}
    \caption{Out-of-domain NLP4SG classification confusion matrix. Most non-NLP4SG papers are properly detected. Comparable amounts of social-good papers are spuriously detected and missed by the classifier compared to the human annotators.}
    \label{tab:domain_transfer}

\end{table}

\begin{figure*}[t!]
    \centering
    \includegraphics[width=\linewidth]{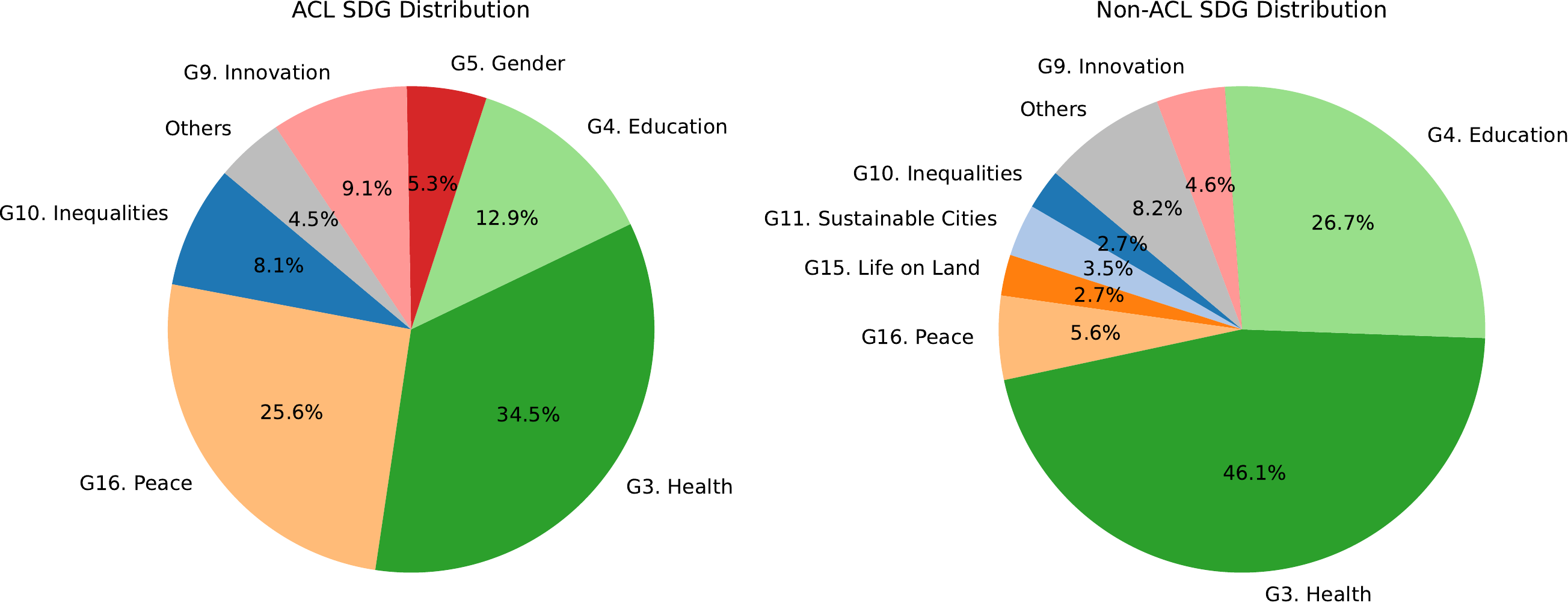}
    \caption{Proportions of SDG topics across venue types. The targets of social good research have different distributions between \textsc{acl}-associated venues and other conferences. \textsc{acl} venues have a greater focus on topics related to peace, innovation, and inequality while \textsc{external} venues have a greater focus on those related to health and education.}
    \label{fig:sdg}
\end{figure*}

\section{Findings}
\label{sec:distributional}

Using this metadata-augmented corpus, we identify large-scale trends in the landscape of NLP4SG. 

\subsection{Distributional Differences by Author and Venue Type}

As the NLP community seeks to interrogate its role in advancing social good concerns, a natural question is where different types of authors engaged with our community choose to publish social good work. Figure \ref{fig:f1} visualizes broad distributional trends in NLP4SG publication patterns, showing the proportion of NLP papers classified as NLP4SG among work using NLP techniques by both \textsc{acl} authors and non-\textsc{acl} authors in \textsc{acl}, \textsc{acl-adjacent}, and \textsc{external} venues.

Our first set of analyses regards author behavior, with a particular focus on \textsc{acl} authors, where we make two key observations. First, papers by \textsc{acl} authors are more likely to address concerns of social good when they appear in \textsc{acl-adjacent} venues rather than core \textsc{acl} venues (one-sided two-proportion z-test, z = 13.5035, p < 0.001). Second, we find that papers by \textsc{acl} authors are also significantly more likely to address social good when they appear in \textsc{external} venues compared to \textsc{acl} venues of any kind (z = 71.4087, p < 0.001); distributionally, we find that for \textsc{acl} authors, the proportion of NLP4SG papers is roughly three times higher outside of \textsc{acl} than within it.

Looking to the behavior of non-\textsc{acl} authors, we find that NLP papers by non-\textsc{acl} authors in \textsc{external} venues represent the category most likely to address social good. This is true both in terms of the proportion of these papers that tackle social good concerns, and dramatically so in the raw number of such papers that appear: the number of NLP4SG papers by non-\textsc{acl} authors in \textsc{external} venues approaches an order of magnitude more than those by \textsc{acl} authors.

Considering change over time (Figure \ref{fig:yearly}), while we observe the same increase in NLP4SG within ACL identified by previous research \cite{adauto-etal-2023-beyond}, but find that this trend is yet stronger in \textsc{acl-adjacent} venues. In \textsc{external} venues the share of NLP4SG papers among NLP papers has always been relatively high but showed a dramatic growth in the mid-2000s and continues to increase in recent years.

\subsection{Topical Differences Across Venue Types}

A natural follow-up question is whether NLP4SG papers that appear in \textsc{acl} venues differ in the types of social good concerns they address compared to those outside it. We use an LLM-based approach to estimate the category of social good for each paper classified as NLP4SG in our corpus. 

As discussed in \citet{adauto-etal-2023-beyond}, InstructGPT \cite{ouyang2022traininglanguagemodelsfollow} demonstrated superior performance in classifying papers according to the UN SDGs. 
These include categories such as ``Health,'' ``Peace,'' and ``Education.''
Building on this, we employ GPT-4o \cite{openai2024gpt4ocard} to classify two sets of NLP4SG papers---those within the \textsc{acl} Anthology (including both \textsc{acl} and \textsc{acl-adjacent} venues) and those outside of it (\textsc{external})---into one of the 17 SDGs.
The prompt template used for classification and the full list of SDGs is provided in Appendix~\ref{sec:prompt_template}. 
We conduct the classification using a greedy decoding strategy.

We note topical differences between social good detected inside and outside of the \textsc{acl} Anthology, illustrated in Figure \ref{fig:sdg}. The \textsc{acl} dataset has a comparatively stronger focus on peace (25.6\%), innovation (9.1\%), and inequalities (8.1\%) compared to the \textsc{external} dataset. The higher proportion of peace-related research in \textsc{acl} may reflect the community’s focus on hate speech detection and related areas that aim to foster peaceful interactions. By contrast, the \textsc{external} dataset has a predominant focus on health (46.5\%) and education (25.1\%), likely reflecting applications of NLP work to targeted practical domains. Overall, despite some similarities, these results suggest that NLP4SG work within ACL and NLP4SG work outside of ACL tend to prioritize different aspects of social good.

\begin{table*}[ht!]
    \centering
    \resizebox{\textwidth}{!}{
    \begin{tabular}{lccc}
        & \multicolumn{1}{c}{\cellcolor{lightgray}\textbf{Total NLP Papers}} & \multicolumn{1}{c}{\cellcolor{lightgray}\textbf{NLP4SG Papers}} & \multicolumn{1}{c}{\cellcolor{lightgray}\textbf{NLP4SG \%}} \\
    \hline
        \multicolumn{1}{|l|}{\textbf{Social Sciences}} & \multicolumn{1}{c|}{6{,}288} & \multicolumn{1}{c|}{3{,}470} & \multicolumn{1}{c|}{55.2\%} \\
    \hline
        \multicolumn{4}{|p{\textwidth}|}{\small \hspace{1em} \textcolor{gray}{\textbf{Top 10 NLP4SG venues:} International Journal of Environmental Research and Public Health, English Language Teaching, SAGE Open, International Journal of Emerging Technologies in Learning (iJET), Cogent Education, Reading and Writing, Scientometrics, Information, Frontiers in Education, Journal of Environmental and Public Health}} \\
    \hline
        \multicolumn{1}{|l|}{\textbf{Health \& Medical Sciences}} & \multicolumn{1}{c|}{26{,}734} &  \multicolumn{1}{c|}{14{,}021} & \multicolumn{1}{c|}{52.4\%} \\
    \hline
        \multicolumn{4}{|p{\textwidth}|}{\small \hspace{1em} \textcolor{gray}{\textbf{Top 10 NLP4SG venues:} PLOS ONE, Frontiers in Psychology, BMC Medical Informatics and Decision Making, International Journal of Environmental Research and Public Health, Frontiers in Genetics, Nucleic Acids Research, Diagnostics, Frontiers in Oncology, JMIR Medical Informatics, Frontiers in Neuroscience}} \\
    \hline
        \multicolumn{1}{|l|}{\textbf{Life Sciences \& Earth Sciences}} & \multicolumn{1}{c|}{26{,}325} & \multicolumn{1}{c|}{11{,}680} & \multicolumn{1}{c|}{44.4\%} \\
    \hline
        \multicolumn{4}{|p{\textwidth}|}{\small \hspace{1em} \textcolor{gray}{\textbf{Top 10 NLP4SG venues:} PLOS ONE, BMC Bioinformatics, Scientific Reports, Bioinformatics, Nucleic Acids Research, Database, PLOS Computational Biology, Sustainability, Heliyon, Frontiers in Plant Science}} \\
    \hline
        \multicolumn{1}{|l|}{\textbf{Humanities, Literature \& Arts}} & \multicolumn{1}{c|}{3{,}104} & \multicolumn{1}{c|}{1{,}281} & \multicolumn{1}{c|}{41.3\%} \\
    \hline
        \multicolumn{4}{|p{\textwidth}|}{\small \hspace{1em} \textcolor{gray}{\textbf{Top 10 NLP4SG venues:} English Language Teaching, Journal of Language and Linguistic Studies, Cogent Arts \& Humanities, Languages, English Education, Studies in Second Language Learning and Teaching, Journal of Child Language, Language Policy, Applied Psycholinguistics, Journal of Psycholinguistic Research}} \\
    \hline
        \multicolumn{1}{|l|}{\textbf{Business, Economics \& Management}} & \multicolumn{1}{c|}{649} & \multicolumn{1}{c|}{239} & \multicolumn{1}{c|}{36.8\%} \\
    \hline
        \multicolumn{4}{|p{\textwidth}|}{\small \hspace{1em} \textcolor{gray}{\textbf{Top 10 NLP4SG venues:} Natural Hazards and Earth System Sciences, Natural Hazards, Ekonomska Istrazivanja-Economic Research, Cogent Business \& Management, International Research Journal of Tamil, Cogent Economics \& Finance, Journal of Artificial Societies and Social Simulation, International Journal of Genomics, Disaster Medicine and Public Health Preparedness, International Journal of Forecasting}} \\
    \hline
        \multicolumn{1}{|l|}{\textbf{Chemical \& Material Sciences}} & \multicolumn{1}{c|}{3{,}173} & \multicolumn{1}{c|}{1{,}020} & \multicolumn{1}{c|}{32.1\%} \\
    \hline
        \multicolumn{4}{|p{\textwidth}|}{\small \hspace{1em} \textcolor{gray}{\textbf{Top 10 NLP4SG venues:} Nucleic Acids Research, Journal of Computer Science, Molecules, Materials, Biomolecules, Molecular \& Cellular Proteomics, Science Advances, Biosensors, RSC Advances, Chemical Science}} \\
    \hline
        \multicolumn{1}{|l|}{\textbf{Engineering \& Computer Science}} & \multicolumn{1}{c|}{41{,}619} & \multicolumn{1}{c|}{13{,}168} & \multicolumn{1}{c|}{31.6\%} \\
    \hline
        \multicolumn{4}{|p{\textwidth}|}{\small \hspace{1em} \textcolor{gray}{\textbf{Top 10 NLP4SG venues:} BMC Bioinformatics, Sensors, IEEE Access, BMC Medical Informatics and Decision Making, Bioinformatics, Applied Sciences, Remote Sensing, Database, JMIR Medical Informatics, PLOS Computational Biology}} \\
    \hline
        \multicolumn{1}{|l|}{\textbf{Physics \& Mathematics}} & \multicolumn{1}{c|}{15{,}618} & \multicolumn{1}{c|}{2{,}825} & \multicolumn{1}{c|}{18.1\%} \\
    \hline
        \multicolumn{4}{|p{\textwidth}|}{\small \hspace{1em} \textcolor{gray}{\textbf{Top 10 NLP4SG venues:} Journal of Physics, Entropy, Mathematics, Symmetry, Physical Review, Chaos Solitons \& Fractals, Discrete Dynamics in Nature and Society, Trends in Hearing, The Annals of Applied Statistics, Statistics in Medicine}} \\
    \hline
        \multicolumn{1}{|l|}{\textbf{Overall}} & \multicolumn{1}{c|}{\textbf{98{,}753}} & \multicolumn{1}{c|}{\textbf{35{,}398}} & \multicolumn{1}{c|}{\textbf{35.8\%}} \\
    \hline 
    \end{tabular}}
    \caption{Disciplines differ in their propensity to use NLP methods to address social good questions. Over half of NLP work in Social Sciences and Health \& Medical Sciences is focused on social good compared to less than 20\% of work in Physics \& Mathematics.}
    \label{tab:nlp4sg_by_category}
\end{table*}

\begin{table*}[h]
    \centering
    \begin{tabular}{lccc}
    \hline
        \textbf{Category} & \textbf{NLP4SG Coefficient} & \textbf{p-value} & \textbf{significance} \\
    \hline
        Business, Economics \& Management     & -2.357  & 0.4336  & \\
        Chemical \& Material Sciences          & 26.199  & 0.0000 & ***\\
        Engineering \& Computer Science        & -24.170 & 0.0000 & ***\\
        Health \& Medical Sciences             & -22.944 & 0.0000 & ***\\
        Humanities, Literature \& Arts         & 2.883   & 0.0000 & ***\\
        Life Sciences \& Earth Sciences        & -19.423 & 0.0000 & ***\\
        Physics \& Mathematics                 & -4.152  & 0.0000 & ***\\
        Social Sciences                       & 19.493  & 0.0000 & ***\\
    \hline
    \end{tabular}
    \caption{Regression results predicting venue-level h5 index from paper NLP4SG  classification across eight coarse disciplinary categories. Broadly speaking, NLP work focused on social good tends to be published in lower-impact venues than other NLP work. However, for certain disciplines like Social Sciences, this trend is reversed.}
    \label{tab:h5index}
\end{table*}


\subsection{Differences by Venue Discipline}
\label{section:venue_discipline}

We leverage Google Scholar metadata and venue-level disciplinary classifications (described in Section \ref{subsection:venue_type}) to examine fine-grained venue-level differences in NLP4SG papers.

One question of interest is whether other computer science venues outside of \textsc{acl} show similar trends to \textsc{acl} regarding the use of NLP methods for social good. To quantify this, we identify CS venues outside of \textsc{acl} as those appearing in one of the following subcategories of `Engineering \& Computer Science': `Artificial Intelligence', `Computational Linguistics', `Computer Graphics', `Computer Vision \& Pattern Recognition', `Computing Systems', `Data Mining \& Analysis', `Databases \& Information Systems', `Human Computer Interaction', `Library \& Information Science', `Robotics', `Signal Processing', `Software Systems', `Theoretical Computer Science'. 

Searching our dataset for these subcategories, we identified a set of 161 CS venues associated with 10,688 papers. Of these, 10\% of the 2,902 \textsc{acl}-authored papers were classified as NLP4SG, compared to 16.9\% of the 7,766 by non-\textsc{acl} authors. These numbers are quite similar to those we observed for our \textsc{acl} and \textsc{acl-adjacent} venue categories (see Figure \ref{fig:f1}), suggesting that \textsc{acl} is comparable to other large computational communities in terms of the proportion of NLP research addressing social good. 

Moving beyond only computer science venues to the full spectrum of disciplines, we aim to ask whether the diverse disciplines that compose our \textsc{external} venues differ in their propensity to use NLP methods to address social good questions.
Table \ref{tab:nlp4sg_by_category} shows the overall distribution of papers identified as corresponding to a known venue in the Google Scholar metadata, classified according to the eight major top-level categories.\footnote{Note that some venues are associated with more than one of the eight categories, so the sum of the individual categories ``total NLP Papers'' exceeds the overall.} We find that disciplinary areas differ in the regularity with which they use NLP methods to address social good concerns. More than half of papers using NLP techniques in venues classified as Social Sciences and Health \& Medical Sciences are identified as NLP4SG, while NLP papers in Physics \& Mathematics address social good less than 20\% of the time.

Lastly, we leverage h5-index metadata from Google Scholar to examine the relationship between NLP4SG work and venue-level impact as measured by citation patterns. The h5-index that we use is defined by Google Scholar as ``the largest number h such that h articles published in 2019-2023 have at least h citations each.''\footnote{\url{https://scholar.google.com/intl/en/scholar/metrics.html}} For each of the eight coarse disciplinary categories, we fit a regression with papers as observations predicting the venue's h5-index against whether a given NLP paper is NLP4SG, and report coefficients of the NLP4SG variable from each regression in Table \ref{tab:h5index}. 

We find interesting discipline-specific patterns. Broadly, NLP4SG work incurs a cost in venue-level impact, such that relative to other papers in a given discipline using NLP methods, NLP4SG papers tend to be published in lower-impact venues. This large-scale pattern mirrors the trend for more NLP4SG in \textsc{acl-adjacent} venues, which tend to be more specialized or regional with higher acceptance rates than \textsc{acl} venues.  However, for a subset of disciplines, particularly in the Chemical \& Material Sciences and Social Sciences, the opposite is true and NLP4SG work is published in higher-impact venues on average relative to other work using NLP methods in that field.

\begin{figure*}[t!]
    \centering
    \includegraphics[width=\linewidth]{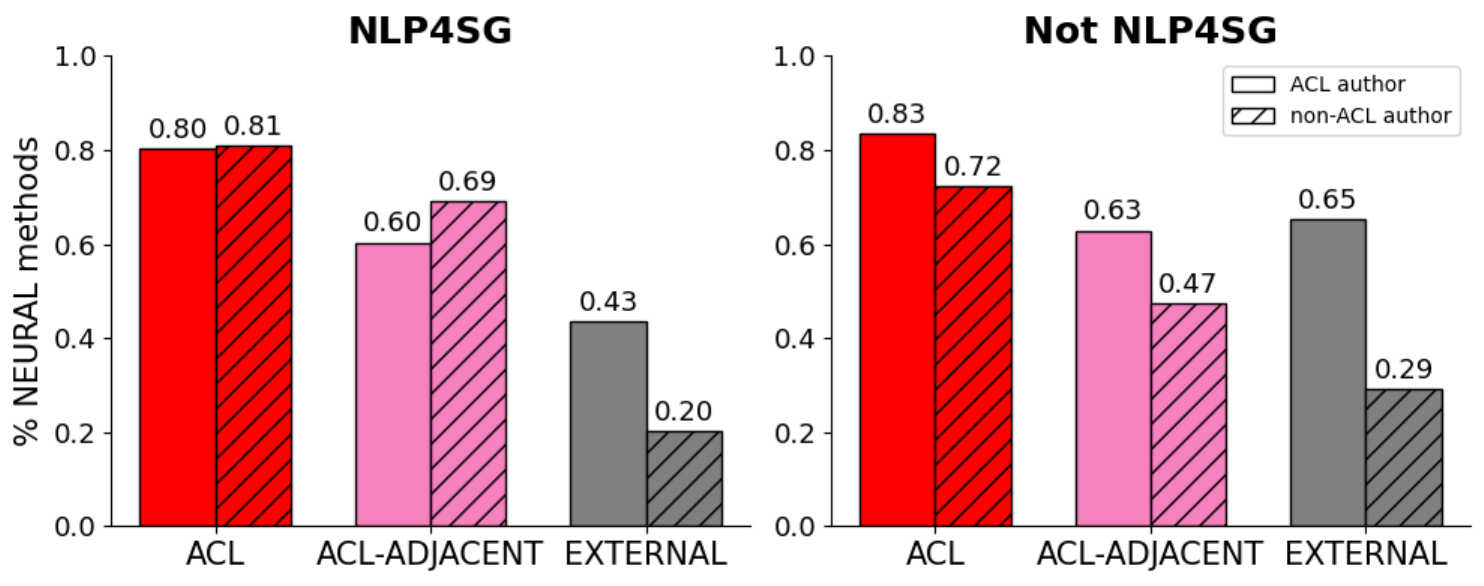}
    \caption{Distribution of neural methods within papers across venue and author types since 2017. Within the \textsc{acl} Anthology, NLP4SG papers are just as likely to rely on neural methods as those not focused on social good. In external venues, NLP4SG papers are relatively less likely to use neural methods.}
    \label{fig:methods}
\end{figure*}

\subsection{Methodological Differences}
Another difference worth exploring involves the NLP methods used in a paper, which may vary significantly by author type or venue type. For example, we find that of all NLP4SG papers since 2020 that explicitly mention LLMs in the abstract, 68.7\% have an \textsc{acl} author.

We explore this possibility of methodological differences by using an LLM to annotate the methodologies used in 20k of the abstracts in our dataset: 10k randomly sampled from \textsc{acl} venues (either core or adjacent), and 10k randomly sampled from \textsc{external} venues. Both random samples were balanced, with 5k NLP4SG and 5k non-NLP4SG papers each. We classify the methods with zero-shot LLM annotation as either neural or traditional in nature. This is of course a highly simplified view into the complex methodological landscape of NLP, which we employ with the goal of understanding high-level trends in how NLP4SG work may or may not make use of the most contemporary methods in the field. Our prompt is provided in Appendix~\ref{sec:prompt_template}. We validated the results with expert human annotations of 50 abstracts, yielding a classification accuracy of 86\%. 

Figure \ref{fig:methods} shows the results for papers published in 2017 or later, grouped by venue type and author type. We see clear differences in methodological focus emerge across these three venue types. Unsurprisingly, core \textsc{acl} venues predominantly rely on neural methods, with approximately 80\% of papers classified as neural. This makes sense given the venue's emphasis on cutting-edge, model-driven methods. Although a majority of papers published at \textsc{acl-adjacent} venues also use neural methods, the overall proportion is somewhat lower. Compared to core \textsc{acl} venues, \textsc{acl-adjacent} venues are often regarded as supporting a broader range of methodologies or placing less emphasis on state-of-the-art results.

While \textsc{external} venues do often use neural methods, non-neural methods are more prevalent. This trend holds across all eight disciplinary categories in the \textsc{external} data except ``Engineering \& Computer Science'' which, much like \textsc{acl}, is predominated by neural methods in the modern era. However, a clear distinction by author type emerges here: when publishing in an \textsc{external} venue, \textsc{acl} authors are using neural methods much more than non-\textsc{acl} authors.

In both core \textsc{acl} and \textsc{acl-adjacent} venues, NLP4SG papers have the same proportion of neural work as all other papers; however, this is not the case for \textsc{external} venues, where papers not focused on social good are more likely to use neural methods than NLP4SG papers across both \textsc{acl} and non-\textsc{acl} author types. 


\section{NLP4SG Within and Beyond ACL}
\label{sec:conclusion}

In the broader context of AI for Social Good~\citep[AI4SG]{tomavsev2020ai}, there has been a concerted effort to apply artificial intelligence methodologies to address societal challenges.
\citet{shi2020artificial} provide an extensive overview of AI4SG applications, discussing various domains where AI has been effectively utilized, such as healthcare~\citep{sarella2024ai}, environmental sustainability~\citep{thulke2024climategpt}, and education~\citep{ferreira2019text}. 
They also identify common challenges in implementing AI solutions for social good, including ethical considerations, data accessibility, and the need for interdisciplinary collaboration.

%

The work that we present here directly builds on \citet{adauto-etal-2023-beyond}, who introduce the NLP4SGPapers dataset and categorize papers based on their relevance to social issues and the UN SDGs. Our results replicate their work: we find a similar level of papers in the \textsc{acl} Anthology classified as NLP4SG (e.g., 17.4\% since 2020).

Yet as we look beyond the confines of \textsc{acl}-associated venues, we see that \textsc{acl} is not where most NLP4SG work is taking place, regardless of who is doing it. Papers by \textsc{acl} authors are dramatically more likely to be classified as NLP4SG when published outside \textsc{acl}, even in the modern era: less than 5\% of all NLP4SG papers since 2020 appear in \textsc{acl} or \textsc{acl-adjacent} venues. The rest are in \textsc{external} venues, the majority of which are written by non-\textsc{acl} authors. Our findings on the distribution of NLP4SG papers further evidence the breadth and depth of NLP as a tool for social good. Indeed, in some disciplines like Social Sciences and Health \& Medical Sciences, when NLP is used (and it is used frequently), it is most often in the service of social good questions.

There are likely substantial differences between NLP4SG work by author and venue that remain to be understood. To better understand the landscape of NLP4SG across various academic venues, future work can conduct comparative analyses that examine how different venues contribute and relate to one another---for example, via citation networks \citep{mosbach2024insights, wahle2023we}, a dynamic we do not address here. Though most NLP4SG work happens outside of \textsc{acl}, it may be that much external work in this area looks to \textsc{acl} as a core source of novel methods. 

To conclude, we return to the agenda-setting questions for the \textsc{acl} community posed in the introduction: 

\begin{quote}
     (1) When \textsc{acl} authors are doing work oriented towards social good, do they send that work to \textsc{acl} venues?

     (2) Is \textsc{acl} the place where most NLP4SG work is taking place? 
\end{quote}

Surprisingly, we find that the answer to both of these questions is a clear ``no''. The evidence strongly suggests that when \textsc{acl}-associated authors seek to publish work targeting questions of social good, they tend to look to venues beyond \textsc{acl}. Moreover, the majority of work using NLP methods for social good happens outside of \textsc{acl}. The reasons for these trends are less clear. One place to look is the NLP Community Metasurvey \cite{michael-etal-2023-nlp}. While this survey did not explicitly ask about social good, the findings suggest that the \textsc{acl} community greatly underestimates its own belief in the value of interdisciplinary science, which includes work addressing social good questions almost by definition. Perhaps researchers feel that NLP4SG work is undervalued in the community, when this may not even be the case.

As a first step, we propose that these findings can serve not as an indictment but rather an inspiration. As \textsc{acl} researchers, we should recognize that many members of our community do publish substantial amounts of NLP4SG work in other venues, and we should continue to advocate that such work be encouraged at *CL conferences by concrete mechanisms such as theme tracks, workshops, and the selection of keynote speakers. A small but important indicator of these ongoing developments in our community is that EMNLP 2025 is the first major NLP conference in which ``NLP for Social Good" appears explicitly in the call for papers as a part of what was previously the ``Computational Social Science and Cultural Analytics" track. 


Ultimately, we hope that these findings help encourage a discussion within the \textsc{acl} community about our role and continued growth in research advancing social good. To this end, we release our augmented metadata for replication and extended work in this area.


\section*{Limitations}

Most importantly, the evidence presented here is associational rather than causal: we do not know, for example, whether \textsc{acl} authors publish relatively more social good work outside of \textsc{acl} because they preferred to do so due to publication incentives, or because that work was first rejected at \textsc{acl} before finding another venue, or some other reason. 

We considered the Semantic Scholar corpus appropriate as a source for broad research publications because it is the largest available corpus of open scientific papers, but beyond that we cannot make any strong claims about the representativeness of our dataset. For example, when linking venues to Google Scholar we find poor representation for the ``Business, Economics \& Management'' category, where several top NLP4SG venues are relatively low-impact. However, we note that perfect representativeness of the full scholarly record would be very challenging to achieve under any definition. We also believe differences in representativeness of the corpus would be very unlikely to change any of the core findings in our work. 

We note that conditioning on \textsc{acl} authorship may overlook newer contributors to the field. However, the ``\textsc{acl} author'' constraint requires only that one author meet the three-paper criteria -- and early-career researchers are typically co-authoring papers with more experienced PIs. In addition, we consider an author an ``\textsc{acl} author'' if they have ever written 3 \textsc{acl} papers. Taken together, we argue that these requirements are largely reasonable, but may mean that we may be failing to consider papers as ``\textsc{acl} authored'' when they ultimately should be if any author later goes on to publish additional papers in the community.

In this work we only looked at NLP4SG as a topic area. The \textsc{acl} community is small relative to all of academic publishing,  so it is logically possible that NLP+X for other X could follow similar trends where the total amount of published work on NLP+X is larger outside of \textsc{acl} than within it. Therefore, our finding about the total quantity of publications cannot tell the whole story; however, we feel it is an important piece of the large-scale context of NLP4SG amidst the other findings presented here. An approach similar to that employed here could be used in future work to examine author- and venue-level differences for other areas of NLP.

As a parting thought, social good in NLP is not limited to just published work, but includes other forms of public engagement that may ultimately be more impactful, such as public applications and tools, community and participatory engagement, educational outreach, and publications intended for broad readership. In this case, we argue that looking at published work provides a feasible and meaningful subset of NLP4SG activity to examine.

\section*{Ethical Considerations}

Social good is a hard-to-define concept. We rely on the existing SDG-based definition, but it is possible alternative definitions---and concordant alternative decisions about values related to what constitutes a social good---could substantially shift the findings reported here.

The NLP4SG classification model from \citet{adauto-etal-2023-beyond} is licensed under Apache 2.  A code assistant was used for visualization iterations.  

\bibliography{custom}

\appendix
\clearpage

\section{Top 10 \textsc{external} Venues}
\label{sec:external-venues}

\begin{itemize}
\itemsep0em 
\item Journal of physics
\item PLOS ONE
\item Scientific Reports
\item Frontiers in Psychology
\item Sensors
\item IEEE Access
\item IOP conference series
\item Lecture Notes in Computer Science
\item BMC Bioinformatics
\item Proceedings of the AAAI Conference on Artificial Intelligence
\end{itemize}

\section{Annotation Instructions}
\label{sec:annotation_instructions}
Annotators were provided with NLP4SG binary classification instructions taken directly from \citet{adauto-etal-2023-beyond}, including their Task 1 decision flowchart (Figure 12 in their paper) and the following text.

\textit{Included:}
Directly related to the high-level definition of SDGs, mentioning, e.g., healthcare; mental health (psychocounseling, hope speech); education; facilitating efficient scientific research (which belongs to Goal 9: Industry, innovation and infrastructure); helping employment (job matching, training job skills); helping collaboration among decision makers.
Related to the fine-grained subcategories of SDGs, e.g., encouraging civic engagement, and enabling social problem tracking for the goal of (Goal 16) Peace, justice and strong institutions.
Social problems in the digital era: e.g., online toxicity, misinformation, privacy protection, and deception detection.

\textit{Excluded:}
General purpose, coarse-grained NLP tasks: machine translation, language modeling, summarization, sentiment analysis, etc.
General purpose, fine-grained NLP tasks: news classification; humor detection; technologies for increasing productivity, e.g., email classification, report generation, meeting note compilation (because they are application-agnostic which could be used for both good and bad purposes, and also a bit too general); textbook-related QA but using it as a benchmark to improve general modeling capabilities; tasks whose data is socially relevant, but the task is neutral (e.g., POS tagging for parliament speech); NLP to help other neutral disciplines, e.g., chemistry; tasks a bit too indirectly related to SDGs, e.g., parsing historical language document, or cultural heritage-related tasks; low resource MT, which bridges resources from one community to another, but is a bit too indirect, and also depends case by case on the actual language community, plus there is a tradeoff between efficiency and equality;
tasks with controversial nature or unknown effect (varying a lot by how people use them in the future): e.g., news comment generation; financial NLP, which could be used in either way to help the economy, or perturb the market for private profits; simulated NLP tools for the battlefield; user-level demographic prediction.

\section{Prompt Templates}
\label{sec:prompt_template}

\begin{tcolorbox}[
    colback=gray!10!white, 
    colframe=gray!50!gray, 
    halign=left, 
    top=0mm, 
    bottom=0mm, 
    left=1mm, 
    right=1mm, 
    boxrule=0.5pt, 
    breakable
]
\footnotesize
\linespread{1}\selectfont
\textbf{Task:} Classify the paper into ONE of the SDG categories.\\
\vspace{5pt}
\textbf{Instructions:}
\begin{itemize}
    \item You will be given a paper title and abstract.
    \item Classify the paper into ONE of the SDG categories.
    \item You should \textbf{ONLY} return ONE SINGLE SDG label, no other text. 
\end{itemize}
\vspace{5pt}

\textbf{Categories with Example Papers:} 

\textbf{G1. Poverty}
\begin{itemize}
    \item Role of AI in poverty alleviation: A bibliometric analysis
\end{itemize}

\textbf{G2. Hunger}
\begin{itemize}
    \item A Gold Standard for CLIR evaluation in the Organic Agriculture Domain
    \item CRITTER: a translation system for agricultural market reports
\end{itemize}

\textbf{G3. Health}
\begin{itemize}
    \item A Treebank for the Healthcare Domain
    \item Automatic Analysis of Patient History Episodes in Bulgarian Hospital Discharge Letters
\end{itemize}

\textbf{G4. Education}
\begin{itemize}
    \item An MT learning environment for computational linguistics students
    \item Salinlahi III: An Intelligent Tutoring System for Filipino Heritage Language Learners
\end{itemize}

\textbf{G5. Gender}
\begin{itemize}
    \item An Annotated Corpus for Sexism Detection in French Tweets
    \item Mitigating Gender Bias in Machine Translation with Target Gender Annotations
\end{itemize}

\textbf{G6. Water}
\begin{itemize}
    \item A conceptual ontology in the water domain of knowledge to bridge the lexical semantics of stratified discursive strata
\end{itemize}

\textbf{G7. Energy}
\begin{itemize}
    \item Artificial intelligence in sustainable energy industry: Status Quo, challenges and opportunities
\end{itemize}

\textbf{G8. Economy}
\begin{itemize}
    \item Multilingual Generation and Summarization of Job Adverts: the TREE Project
    \item Situational Language Training for Hotel Receptionists
\end{itemize}

\textbf{G9. Innovation}
\begin{itemize}
    \item An Annotated Corpus for Machine Reading of Instructions in Wet Lab Protocols
    \item Retrieval of Research-level Mathematical Information Needs: A Test Collection and Technical Terminology Experiment
\end{itemize}

\textbf{G10. Inequalities}
\begin{itemize}
    \item Analyzing Stereotypes in Generative Text Inference Tasks
    \item Recognition of Static Features in Sign Language Using Key-Points
\end{itemize}

\textbf{G11. Sustainable Cities}
\begin{itemize}
    \item FloDusTA: Saudi Tweets Dataset for Flood, Dust Storm, and Traffic Accident Events
    \item Trouble on the Road: Finding Reasons for Commuter Stress from Tweets
\end{itemize}

\textbf{G12. Consumption}
\begin{itemize}
    \item Multiple Teacher Distillation for Robust and Greener Models
\end{itemize}

\textbf{G13. Climate}
\begin{itemize}
    \item CLIMATE-FEVER: A Dataset for Verification of Real-World Climate Claims
    \item Tackling Climate Change with Machine Learning
\end{itemize}

\textbf{G14. Life Below Water}
\begin{itemize}
    \item Marine Variable Linker: Exploring Relations between Changing Variables in Marine Science Literature
    \item Literature-based discovery for Oceanographic climate science
\end{itemize}

\textbf{G15. Life on Land}
\begin{itemize}
    \item Harnessing Artificial Intelligence for Wildlife Conservation
\end{itemize}

\textbf{G16. Peace}
\begin{itemize}
    \item On Unifying Misinformation Detection
    \item Fully Connected Neural Network with Advanced Preprocessor to Identify Aggression on Social Media
\end{itemize}

\textbf{G17. Partnership}
\begin{itemize}
    \item MEDAR: Collaboration between European and Mediterranean Arabic Partners to Support the Development of Language Technology for Arabic
    \item The Telling Tail: Signals of Success in Electronic Negotiation Texts
\end{itemize}

\vspace{5pt}
\textbf{Paper Title:} \{paper-title\} \\
\textbf{Paper Abstract:} \{paper-abstract\}
\end{tcolorbox}

\begin{tcolorbox}[
    colback=gray!10!white, 
    colframe=gray!50!gray, 
    halign=left, 
    top=0mm, 
    bottom=0mm, 
    left=1mm, 
    right=1mm, 
    boxrule=0.5pt,     
]
\footnotesize
\linespread{1}\selectfont
\textbf{Instructions} Given the below academic paper's abstract, extract the primary methodological paradigm applied by the researchers.
\\
\vspace{5pt}
\textbf{text}: \{abstract\}
\vspace{5pt} 
\begin{itemize}
    \item Examples of methods in the NEURAL category are large language models (LLMs), generative AI, deep learning, neural networks, LSTMs, vector representations, and embeddings.
    \item Examples of methods in the TRADITIONAL category are statistical NLP, logistic regression, support vector machines, random forests, structure prediction, lexicons, regular expressions, and rule-based heuristics.
    \item The possible two responses for [methodology] are: NEURAL TRADITIONAL
\end{itemize}
    
Return only: [methodology]

Do NOT return anything other than NEURAL or TRADITIONAL.
\end{tcolorbox}

\end{document}